\documentclass[sigconf]{acmart}
\usepackage[autostyle]{csquotes}  
\AtBeginDocument{%
  \providecommand\BibTeX{{%
    \normalfont B\kern-0.5em{\scshape i\kern-0.25em b}\kern-0.8em\TeX}}}

\newcommand{\voc}[2]{\texttt{#1:\allowbreak #2}}
\newcommand{\kg}{\textit{Quote\-KG}}

\setcopyright{none}
\renewcommand\footnotetextcopyrightpermission[1]{}
\usepackage{enumitem}

\begin{document}

\title{Comprehensive Event Representations using Event Knowledge Graphs and Natural Language Processing}

\author{Tin Kuculo}
\affiliation{Supervised by Prof. Dr. Wolfgang Nejdl, Prof. Dr. Elena Demidova and Dr. Simon Gottschalk
  \institution{L3S Research Center, Leibniz Universität Hannover}
  \city{Hannover}
  \country{Germany}}
\email{kuculo@L3S.de}

\renewcommand{\shortauthors}{Tin Kuculo}

\begin{abstract}
Recent work has utilised knowledge-aware approaches to natural language understanding, question answering, recommendation systems, and other tasks. These approaches rely on well-constructed and large-scale knowledge graphs that can be useful for many downstream applications and empower knowledge-aware models with commonsense reasoning. Such knowledge graphs are constructed through knowledge acquisition tasks such as relation extraction and knowledge graph completion. This work seeks to utilise and build on the growing body of work that uses findings from the field of natural language processing (NLP) to extract knowledge from text and build knowledge graphs. The focus of this research project is on how we can use transformer-based approaches to extract and contextualise event information, matching it to existing ontologies, to build a comprehensive knowledge of graph-based event representations. Specifically, sub-event extraction is used as a way of creating sub-event-aware event representations. These event representations are then further enriched through fine-grained location extraction and contextualised through the alignment of historically relevant quotes.


\end{abstract}



\maketitle
\pagestyle{plain}

{\let\thefootnote\relax\footnotetext{{© {Tin Kuculo} {2022}. This is the author's version of the work. It is posted here for your personal use. Not for redistribution. The definitive Version of Record was published in {Companion Proceedings of the Web Conference 2022}, \url{https://doi.org/10.1145/3487553.3524199}.}}}


\section{Introduction}

Knowledge graphs, structured representations of facts comprising entities, relationships, and semantic descriptions, have recently gained both academic and industrial attention \cite{ji2021survey}. By creating comprehensive machine-readable knowledge graphs about the world's entities, their semantic properties, and their relationships, remarkable possibilities open up. Potential applications include the creation of machine-readable encyclopedias that can be queried with high precision, similar to semantic databases; quick and accurate mapping of textual phrases to entities in the knowledge graph; an ability to answer natural language questions about entities such as who, where, when. Finally, world knowledge is a crucial component in providing disambiguating context for machine translation and can be a catalyst for the acquisition of further knowledge, as well as for the growth and maintenance of the knowledge graph \cite{weikum2010information}. 

Knowledge graphs mainly cover static information such as a person's date of birth and birthplace, illustrating the state of the world rather than changes over time. However, considering the dynamic nature of events can be very important for the reconstruction of someone's past or the extensive past of local industries, regions, or organisations. It is, therefore, necessary to build a different type of structured database based on events rather than entities and entity-based real facts. Capturing this dynamic knowledge requires considering events as the unit for storing knowledge \cite{rospocher2016building}. This need has resulted in the study and creation of event knowledge graphs. 

There are two different approaches to understanding and processing events, one more from the Semantic Web and one more from the NLP perspective. In knowledge graphs such as DBpedia and Wikidata, typically, events are clearly distinguished and named events such as Brexit, FIFA World Cup 2014. This is often seen in knowledge graphs. The second view is that of fine-grained events found in text, often as sub-events in a bigger context. For example, a text about Brexit might cover the visit of the UK's prime minister Theresa May in Berlin: this is the approach often used in NLP. In my work, I want to bring these two approaches together. For example, consider a properly modelled \textit{Visit} event, which is linked to the \textit{Brexit} event in a knowledge graph. To extract and model this kind of information, we use Wikipedia as our source of unstructured information and Wikidata to structure the outputs of our NLP components.  

In this work, by utilising event extraction, we find new events and form their representations. Event representations can still benefit from additional contextualisation. We provide this additional context in two ways:

(i) Quotes: while static information such as the time and location of a given event is understood across languages, the impact of a given event on specific populations varies across linguistic and cultural contexts of those populations. In our work, we argue that this impact can be captured and analysed across time through the lens of quotes. Specifically, the quotes said by persons of public interest are often relevant from the historical perspective.
Examples include \textit{"Tear down this wall"} by the United States President Ronald Reagan in the context of the fall of the Berlin Wall.

(ii) Fine-grained locations: another perspective is the provision of locations with higher precision. While knowledge graphs often provide locations of events, they are often not fine-grained. For example, for an event such as the 2022 Winter Olympics, knowledge graphs might only contain the country, China, or the city, i.e., Beijing, but not the actual competition sites. In this example, we want to do better and find the exact venues, such as the Beijing Wukesong Sports Center.

\section{Problem}

Given a set of named events, we want to extract all relevant knowledge about them and contextually interlink them to persons and their quotes. This is accomplished through three separate components.
\begin{itemize}
\item Sub-event Extraction: A processing pipeline is created such that for each event, all of its sub-events are extracted and linked with its entity arguments to corresponding Wikidata objects and classes. 
\item Fine-Grained Location Extraction: For additional contextualization of events, OpenStreetMap is used to link geographic entities present in the text of an event's Wikipedia page. Through a representation learning approach, correct geographic entities are then selected as the most precise available fine-grained location information about the event.
\item Quote Extraction and Cross-lingual Alignment: The third processing pipeline utilizes Wikiquote – an online collection of quotes\footnote{\url{https://en.wikiquote.org/wiki/Main_Page}}. Using Wikiquote as a data source, we focus on Wikiquote pages about persons. The quotes by these persons are then extracted and interlinked with their contexts. Finally, encoding the quotes using a cross-lingual language model, we cluster them together across languages.  
\end{itemize}

\section{Proposed approach}

\begin{figure*}[t]
    \centering
    \includegraphics{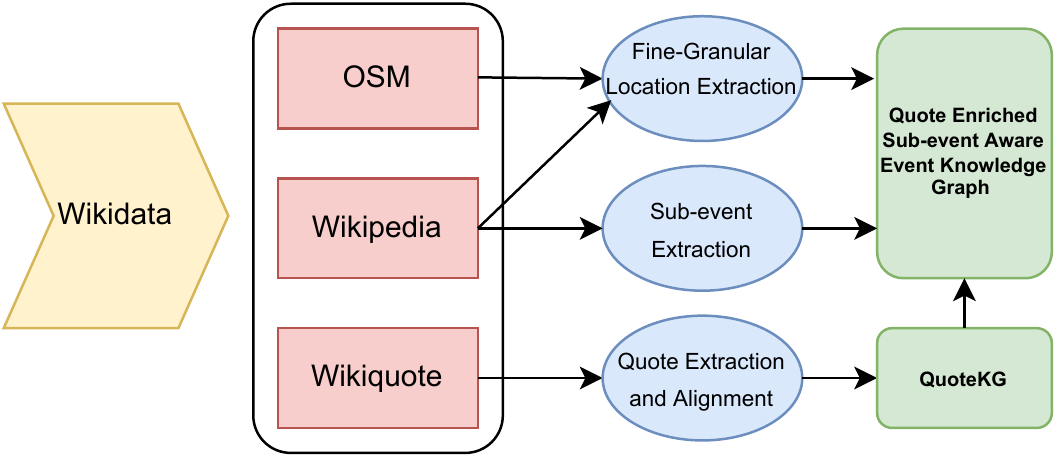}
    \caption{General structure of the research project.}
    \label{fig:structure}
\end{figure*}

In this section, we briefly present our approach to the specific tasks we focus on. These tasks are event extraction, and event enrichment and contextualization. The general approach can be seen in Figure ~\ref{fig:structure}.

For the event extraction, a set of events of interest is defined. For the selected events, we take their corresponding Wikipedia pages and run our sub-event extraction pipeline. The pipeline consists of three components: the event type extraction, the argument extraction component, and the knowledge modelling component that matches our extracted information to the Wikidata ontology. 

To enrich and contextualize our events, we take on two tasks. In the first, we address the extraction of fine-grained locations. 
While some events' locations can be accurately described by the names of countries or cities they have occurred in, others would benefit from more fine-grained location information, e.g., the specific building in which the event occurred instead of the city. To increase the precision of event location information, we take a mixed approach of relation extraction, entity linking, and link prediction. Lastly, for the contextualization of events, we extract quotes of public figures and interlink them with event and entity information.

\section{Methodology}

In this section, the main components of the research project will be described—namely, the event extraction and the contextualisation and enrichment of events.

\subsection{Event Extraction}
In the first step, a set of events is selected. For each of these events, we take their corresponding Wikipedia page and run an event extraction model \cite{lu2021text2event}. The event extraction component runs on each paragraph of the Wikipedia page and outputs a set of sub-events found in the text. For example, consider the following excerpt in Figure ~\ref{fig:wiki_sentence1}:

    \begin{figure}[H]
    \begin{quotation}
As of 20 January 2022, at least 1,488 protesters and bystanders, have been shot and killed by police forces and at least 8,702 people detained. \end{quotation}
    \caption[]{Taken from the English Wikipedia page on the Myanmar Protests and modified for simplicity.\protect\footnotemark}
    \label{fig:wiki_sentence1}
    \end{figure}
    
\footnotetext{\url{https://en.wikipedia.org/wiki/2021-2022_Myanmar_protests}}

The event extraction component identifies two sub-events: a \textit{Die} sub-event and an \textit{Arrest} sub-event. In the following step, these are mapped to Wikidata classes \textit{death (Q4)}\footnote{\url{https://www.wikidata.org/wiki/Q4}} and \textit{legal action (Q2135540)}. We extract all their subclasses to be candidate Wikidata classes representing our subevents. Among the subclasses of \textit{death (Q4)} there exists a Wikidata class \textit{extra-judicial killing (Q2717573)}, similarly among the subclasses of \textit{legal action (Q2135540)} there is a \textit{detention (Q1071447)}. For each of the candidate subclasses, we generate a list of questions to be asked by the question answering model \cite{lan2019albert}. According to the returned answers, we select the correct Wikidata class to be assigned as the sub-event type.

To evaluate the performance of this processing pipeline, we generate a test dataset based on existing sub-events in Wikipedia texts. Specifically, here we are referring to sub-events already defined in Wikidata.

    \begin{figure}
    \begin{quotation}
On July 19, the second night of the \textit{2016 Republican National Convention}, Pence won the Republican vice presidential nomination by acclamation. \end{quotation}
    \caption{Example sentence in the Wikipedia article on the 2016 Republican Party vice-presidential candidate selection. Links are italicised.}
    \label{fig:wiki_sentence2}
    \end{figure}

For example, in the sentence shown in Figure~\ref{fig:wiki_sentence2}, the 2016 Republican National Convention is an existing event in Wikidata, with properties such as location, country, point in time. In addition to this automatically generated test dataset based on Wikipedia links, we also plan to manually annotate a selected subset of Wikipedia event articles. To evaluate, we compare our model's extracted properties with the ones present in Wikidata. This evaluation procedure is done in tandem on different baselines. Firstly, to validate the need for the question answering component, we ablate the question answering part of the pipeline and match ACE ontology events directly to Wikidata. This is followed by an evaluation of a SOTA triple extraction model on existing sub-events. Finally, we compare our full approach to an approach that extracts entities and does triple extraction on text.

\subsection{Event Contextualization and Enrichment}
To further enhance our event representations, we look to two directions: increasing the precision of our location information and contextualising events through related quotes of persons of public interest.
\subsubsection{Fine-Grained Location Extraction}
To further enrich event representations, a processing pipeline is used to fill empty location nodes and retrieve more precise location information for existing ones.
Specifically, we use a combined approach of link prediction and location extraction. A multitask framework \cite{stoica2020improving} is used to train a model to jointly predict locations based on existing nodes in EventKG \cite{gottschalk2018eventkg} and extract fine-grained location information from Wikipedia texts of corresponding events. The fine-grained location information is determined via entity-linking to OpenStreetMap (OSM) using WorldKG \cite{dsouza2021worldkg}, and a relation extraction model trained to extract candidate locations and rank them to determine the correct location.

\subsubsection{Quote Extraction and Cross-lingual Alignment}
To support the research of events through the lens of impactful quotes by public figures, we align quotes across languages and enrich them with context, creating the first multilingual knowledge graph of quotes, \kg{} \cite{kuculo2022quotekg}. \kg{} is based on Wikiquote, another wiki-based project. Like Wikipedia, Wikiquote has independent versions for different languages. It consists of pages, each of which covers a particular topic and is divided into subsections and sections. For \kg{}, we focus on Wikiquote pages about persons that contain quotes attributed to them. 

\begin{figure*}[ht]
    \centering
    \includegraphics[width=0.7\textwidth]{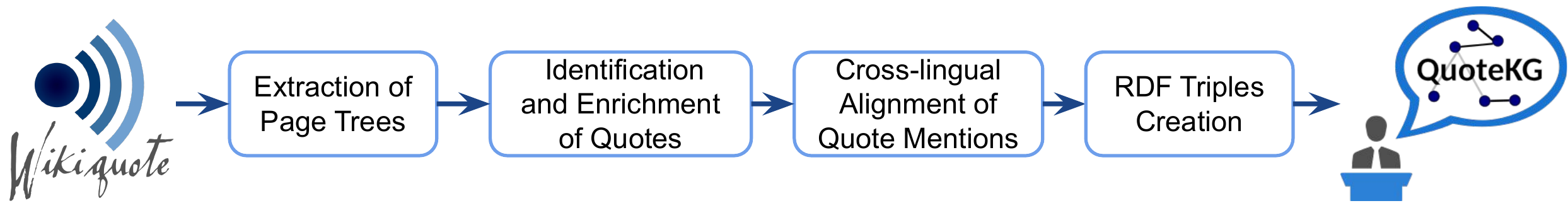}
    \caption{Pipeline to create \kg{} from Wikiquote.}
    \label{fig:qkg_pipeline}
\end{figure*}

\kg{}'s creation pipeline, shown in Figure ~\ref{fig:qkg_pipeline}, begins by processing Wikiquote in all available languages and selecting the pages about specific persons. 

After we identify and enrich the quotes, we need to detect which of them represent mentions of the same quote said by a person of public interest. This task of cross-lingual alignment of quote mentions is treated as a clustering task at the end of which each cluster represents a quote with a set of mentions.

The clustering task is performed for each individual person separately. Given a person's quote mention in a set of languages, the aim is to create clusters of highly similar mentions. To derive a similarity between two mentions, potentially from different languages, we compute the cosine similarity of sentence embeddings derived from the mentions' texts. As an embedding model, a language-agnostic transformer model pre-trained on millions of multilingual paraphrase examples is used, namely, XLM-RoBERTa \cite{conneau2019unsupervised}.
Given these embeddings and the cosine similarity function, clustering is performed by detecting communities of quotes using a nearest-neighbour search. To do so, we utilise UKPLab's Fast Clustering algorithm\footnote{\url{https://github.com/UKPLab/sentence-transformers/blob/master/examples/applications/clustering/}}  as it is optimised for efficient similarity computations of our embeddings.

\section{Preliminary results}

So far, we have created the \kg{} schema based on schema.org as well as a pipeline that extracts quotes from the Wikiquote corpus and aligns them across languages. The \kg{}'s schema was made to model quotes, their relationships with persons and other entities, as well as their different mentions, e.g., translations, typically in different contexts. To this end, \kg{} is based on an extension of the \textit{schema.org} vocabulary that provides a \voc{so}{Quotation}\footnote{Prefix "\voc{so}" stands for \url{https://schema.org/}} class which is re-used.  \kg{} has been made publicly available\footnote{\url{https://quotekg.l3s.uni-hannover.de/}} and includes nearly one million quotes in $52$ languages, said by nearly $69,000$ people of public interest. Table~\ref{tab:statistics} provides detailed statistics for selected languages. 

For the sub-event extraction framework, the question answering pipeline has been completed. We have established the mapping between ACE event types and Wikidata classes. In addition, the entity linking and the question answering part of the pipeline has been built as well. So far, we have identified many cases of potential new sub-events, for example, consider the following sentence taken from the Wikipedia article on \textit{Battle of Latakia}:

\blockquote{the Israeli ships engaged a 560-ton Syrian T43-class minesweeper}
While a sub-event extraction model might detect an \textit{Attack} sub-event, our pipeline detects \textit{naval battle (Q1261499)}. 

Lastly, although the fine-grained location extraction is still in an early stage of development, we find the research well-motivated. After our data collection, we have found $700,000$ events, of which $500,000$  have locations, but only $50,000$ have coordinates.

\begin{table}[t]
\footnotesize
\centering
\caption{Statistics of selected languages in \kg{}.}
\label{tab:statistics}
\begin{tabular}{lrrrr} \toprule
\multicolumn{1}{c}{\textbf{Language}} & \multicolumn{1}{c}{\textbf{Persons}} & \multicolumn{1}{c}{\textbf{Quotes}} & \multicolumn{1}{c}{\textbf{Mentions}} & \multicolumn{1}{c}{\textbf{Mentions with Contexts}}
 \\ \midrule
English & 19,076 & 267,994 & 271,879 & 193,848 \\
Italian & 18,795 & 145,258 & 146,128  & 48,107 \\
German & 3,463 & 16,017 & 16,454  & 4,330 \\
Croatian & 2,720 & 11,019 & 12,944 & 2,045 \\
Welsh & 241 & 457 & 503 & 247 \\ \midrule
\textbf{All Languages} & 69,539 & 882,969 & 964,009 & 412,801 \\ \bottomrule
\end{tabular}
\end{table}

\section{State of the art}

In this section, the latest research in event extraction is addressed, prominent event knowledge graphs, and work done on geolocation. This is followed by an overview of other knowledge graphs containing quotes.

\subsection{Event Knowledge Representation} 
Event knowledge graphs have come about as a way to address the dynamic nature of the real world and as a way to capture the temporal dimension of relations. This is shown in initial work on temporal knowledge graphs such as the Integrated Conflict Early Warning System (ICEWS) dataset and the Global Database of Events, Language and Tone (GDELT). Recently, more work has been done in the financial domain and work motivated by the news domain \cite{charlotte2019searching, al2020lifting}. Work that steps out of domain-specific focus \cite{gottschalk2018eventkg} focuses on the big-picture event properties such as the time, location, and participants of a particular event, not delving into the specific sub-events of an event. One notable exception is the NewsReader project \cite{rospocher2016building} where the authors focus on news articles and use a synset detection based approach, whereas, in my work, the focus is creating enriched representations of important events that contain sub-events as they relate to them. 

\subsection{Event Extraction}
Event extraction, a challenging research task in information extraction, is at the heart of information extraction research. 
Event extraction effectively converts unstructured plain texts into structured event information that describes "who, when, where, what, why" and "how" of real-world events. Traditional event extraction methods mainly relied on complex feature engineering and faced error propagation issues. Some of these issues are ameliorated using deep learning approaches. The early deep learning approaches \cite{chen2015event, nguyen2015event}  used Convolutional Neural Networks (CNN) to automatically extract lexical and sentence-level features. Other research explores RNN-based architectures. JRNN \cite{nguyen2016joint} is proposed with a bidirectional RNN for event extraction. While previous approaches relied heavily on language-specific knowledge and existing NLP tools, \cite{feng2018language} show that learning automatically from data can result in accurate multilingual event detection. 

Sentence-level sequential modelling suffers from the low efficiency in capturing very long-range dependencies. The modelling of structural information in the attention mechanism has gradually attracted the attention of researchers. The JMEE \cite{liu2018jointly} uses a graph convolution network based on attention to jointly model the graph information to extract multiple event triggers and arguments. \cite{nguyen2018graph} investigate a CNN based on dependency trees to perform event detection and are the first to integrate syntax into neural event detection. The network explicitly models the information from entity mentions to improve performance for event detection. 

Since the arrival of transformer models, research has shown its efficacy on event extraction \cite{zhang2019joint,yang2019exploring} . \cite{wadden2019entity} is a BERT-based framework that models text spans and captures within-sentence and cross-sentence context. Similar to our own work, \cite{du2020event,li2020event,liu2020event} frame event extraction as a question answering problem.

\subsection{Event Geolocation}
The task of event geolocation has most recently been studied as a part of event detection in social media analysis  \cite{mousset2020end,middleton2018location}. Other research focuses on Geolocating entities in knowledge graphs \cite{qiu2019knowledge} or geotagging text content in general \cite{kordopatis2017geotagging,avvenuti2018gsp,laparra2020dataset}.
On the other hand, work addressing geotagging of events \cite{halterman2019geolocating, dewandaru2020event} tends to focus on a specific domain. While there is work focusing on fine-grained location estimation \cite{lautenschlager2017statistical}, to my best knowledge, this has not been applied specifically to events and OpenStreetMap geographic objects.

\subsection{Quotes in Knowledge Graphs} DBQuote \cite{piao2015dbquote} is a system for collection and semantic annotation of quotes extracted from Twitter and Wikiquote; however, it covers only two languages (English and Korean) and has not been made available. Other work  \cite{newell2018quote} focuses on English news articles, quotes, and topics to create a database intended for journalistic research. Event-centric knowledge graphs, such as EventKG \cite{gottschalk2019eventkg}, offer insights into human history and events of importance, but they lack quotations that provide context to public figures and their actions. As a consequence, quotes have only had insufficient coverage in the Semantic Web: for example, Wikidata \cite{vrandevcic2014wikidata} has fewer than $400$ instances of the class ``Phrase''\footnote{\url{https://www.wikidata.org/wiki/Q187931}} that are attributed to an author or creator -- most of them only consisting of few words (e.g., ``cogito ergo sum'' and ``covfefe''). To the best of my knowledge, our knowledge graph of quotes, \kg{}, is the first publicly available knowledge graph of quotes.

\section{Conclusions}

In this paper, I have described my research project and its focus on the creation of quote enriched sub-event aware event representations with very fine-grained location information. I have defined the framework to be used for this purpose and have shown some preliminary results. These include \kg{}, a multilingual knowledge graph of quotes, the state of event extraction framework, and the statistics motivating my work on fine-grained event extraction.

\begin{acks}
I thank my supervisors, Prof. Dr. tech. Wolfgang Nejdl, Prof. Dr. Elena Demidova, and Dr. Simon Gottschalk for their feedback and support. This work has been funded by European Union’s Horizon 2020 research and innovation programme under the Marie Skłodowska-Curie grant agreement no. 812997 (Cleopatra).
\end{acks}

\bibliographystyle{ACM-Reference-Format}
\bibliography{bibliography}

\end{document}